\documentclass{article} 
\usepackage[preprint]{colm2025_conference}

\usepackage{microtype}
\usepackage{hyperref}
\usepackage{url}
\usepackage{booktabs}

\usepackage{amsmath}
\usepackage{amssymb}

\usepackage{relsize}
\usepackage{subcaption}
\usepackage{makecell}
\usepackage{multirow}
\usepackage{graphicx}
\usepackage{comment}
\usepackage{xcolor}
\definecolor{promptgreen}{HTML}{13765a}
\definecolor{promptred}{HTML}{c04d5a}

\usepackage{lineno}

\definecolor{darkblue}{rgb}{0, 0, 0.5}
\hypersetup{colorlinks=true, citecolor=darkblue, linkcolor=darkblue, urlcolor=darkblue}

\newcommand{\modelname}{\emph{Nanda}}
\newcommand{\modelnamefull}{\emph{Llama-3-Nanda-10B-Chat}}
\newcommand{\basemodelnamefull}{\emph{Llama-3-8B-Instruct}}
\newcommand{\basemodelnameshort}{\emph{Llama-3-Instruct}}

\newcommand{\jais}{\emph{Jais}}

\title{Llama-3-Nanda-10B-Chat:\\ An Open Generative Large Language Model for Hindi}

\author{Monojit Choudhury$^{1}$, Shivam Chauhan$^{1}$, Rocktim Jyoti Das$^{1}$, Dhruv Sahnan$^{1}$, 
\\
\textbf{Xudong Han$^{1}$, Haonan Li$^{1}$, Aaryamonvikram Singh$^{1}$, Alok Anil Jadhav$^{1}$, }
\\
\textbf{Utkarsh Agarwal$^{1}$, Mukund Choudhary$^{1}$, Debopriyo Banerjee$^{1}$, Fajri Koto$^{1}$, }
\\
\textbf{Junaid Bhat$^{2}$, Awantika Shukla$^{2}$, Samujjwal Ghosh$^{2}$, Samta Kamboj$^{2}$, Onkar Pandit$^{2}$, }
\\
\textbf{Lalit Pradhan$^{2}$, Rahul Pal$^{2}$, Sunil Sahu$^{2}$, Soundar Doraiswamy$^{2}$, Parvez Mullah$^{2}$,} 
\\
\textbf{Ali El Filali$^{2}$, Neha Sengupta$^{2}$, Gokul Ramakrishnan$^{3}$, Rituraj Joshi$^{3}$, Gurpreet Gosal$^{3}$, }
\\
\textbf{Avraham Sheinin$^{3}$, Natalia Vassilieva$^{3}$, Preslav Nakov$^{1}$}
\\
\\
$^1$ Mohamed Bin Zayed University of Artificial Intelligence, UAE\\
$^2$ Inception, UAE\\
$^3$ Cerebras Systems\\
}

\begin{document}

\ifcolmsubmission
\linenumbers
\fi

\maketitle

\begin{abstract}

    Developing high-quality large language models (LLMs) for moderately resourced languages presents unique challenges in data availability, model adaptation, and evaluation. We introduce \modelnamefull{}, or \modelname{} for short, a state-of-the-art Hindi-centric instruction-tuned generative LLM, designed to push the boundaries of open-source Hindi language models. Built upon Llama-3-8B, \modelname{} incorporates continuous pretraining with expanded transformer blocks, leveraging the Llama Pro methodology. A key challenge was the limited availability of high-quality Hindi text data; we addressed this through rigorous data curation, augmentation, and strategic bilingual training, balancing Hindi and English corpora to optimize cross-linguistic knowledge transfer. With 10 billion parameters, \modelname{} stands among the top-performing open-source Hindi and multilingual models of similar scale, demonstrating significant advantages over many existing models. We provide an in-depth discussion of training strategies, fine-tuning techniques, safety alignment, and evaluation metrics, demonstrating how these approaches enabled \modelname{} to achieve state-of-the-art results. By open-sourcing \modelname{}, we aim to advance research in Hindi LLMs and support a wide range of real-world applications across academia, industry, and public services.
\end{abstract}

\section{Introduction}
\label{sec:introduction}

Recent advances in transformer-based large language models (LLMs) have significantly transformed natural language processing (NLP), enabling impressive reasoning and instruction-following capabilities. However, most development has remained English-centric. While multilingual LLMs such as Falcon~\citep{falcon40b}, PaLM~\citep{chowdhery2022palm}, Aya~\citep{Ustun2024AyaMA}, and Llama-3.1~\citep{Dubey2024TheL3} attempt to broaden linguistic coverage, their pretraining continues to rely heavily on English-dominated corpora, limiting their performance in underrepresented languages. Additionally, these models often suffer from the \textit{``curse of multilinguality''}~\citep{Pfeiffer2022LiftingTC, Arivazhagan2019MassivelyMN, conneau2020}, where performance degrades as the number of supported languages increases. To address this, we introduce \modelnamefull{} (\modelname{}), a 10B-parameter decoder-only LLM tailored for Hindi, the fourth most spoken language globally.\footnote{\url{https://en.wikipedia.org/wiki/Hindi}}$^,$\footnote{\url{https://www.worlddata.info/languages/hindi.php}} We present \modelname{}'s model card in Appendix \ref{sec:app:model_card}.

Building high-quality Hindi LLMs presents challenges due to limited data availability~\citep{joshi-etal-2020-state}. In contrast to English, which benefits from corpora of up to 15 trillion tokens~\citep{txt360data2024}, Hindi resources are scarce. To mitigate this, we curated a 65B token Hindi corpus for continued pretraining and developed a data processing pipeline to ensure high-quality data, which includes code mixed (with English) and romanized Hindi examples. Unlike massively multilingual models such as Bloom~\citep{Scao2022BloomA1} and Aya~\citep{Ustun2024AyaMA},
\modelname{} focuses solely on Hindi and English. 
We use an equal ratio of Hindi-English tokens for pretraining and apply oversampling during instruction-tuning to balance the 64.5M English and 43.5M Hindi tokens in the instruction-tuning dataset.

\modelname{} builds on Llama-3~\citep{Dubey2024TheL3}, and incorporates recent breakthroughs such as RoPE~\citep{Su2021RoFormerET} and grouped-query attention~\citep{Ainslie2023GQATG}, along with a custom-built tokenizer for bilingual optimization. We evaluate the model across Hindi and English benchmarks in reasoning, factuality, safety, bias and generation. Results show that \modelname{} is one of the best-performing Hindi-English bilingual language model, achieving competitive results in reasoning and factuality tasks while outperforming similarly sized models in text generation. These results mark a promising step toward robust, high-quality LLMs for Indic languages.

\section{Pretraining Data Preparation}
\label{sec:pretraining_data}
\modelname{} is pre-trained on billions of words to build a strong foundation in Hindi, with a knowledge base tailored to language's cultural nuances.
We curated a large pre-training dataset by incorporating diverse Hindi-language sources, including websites, news articles, books, and , Wikipedia.
This dataset integrates resources such as Hindi-specific datasets from HuggingFace, IIT-Bombay English-Hindi Parallel Corpus~\citep{kunchukuttan-etal-2018-iit} and High Performance Language Technologies' Multilingual Datasets~\citep{burchell2025expandedmassivemultilingualdataset}.
In total, our pre-processed dataset comprises of \textbf{65 billion tokens of Hindi data}.


\subsection{Preprocessing Pipeline}
\label{sec:preprocess_pipeline}
\begin{figure}[t]
    \centering
    \resizebox{0.75\textwidth}{!}{
    \includegraphics{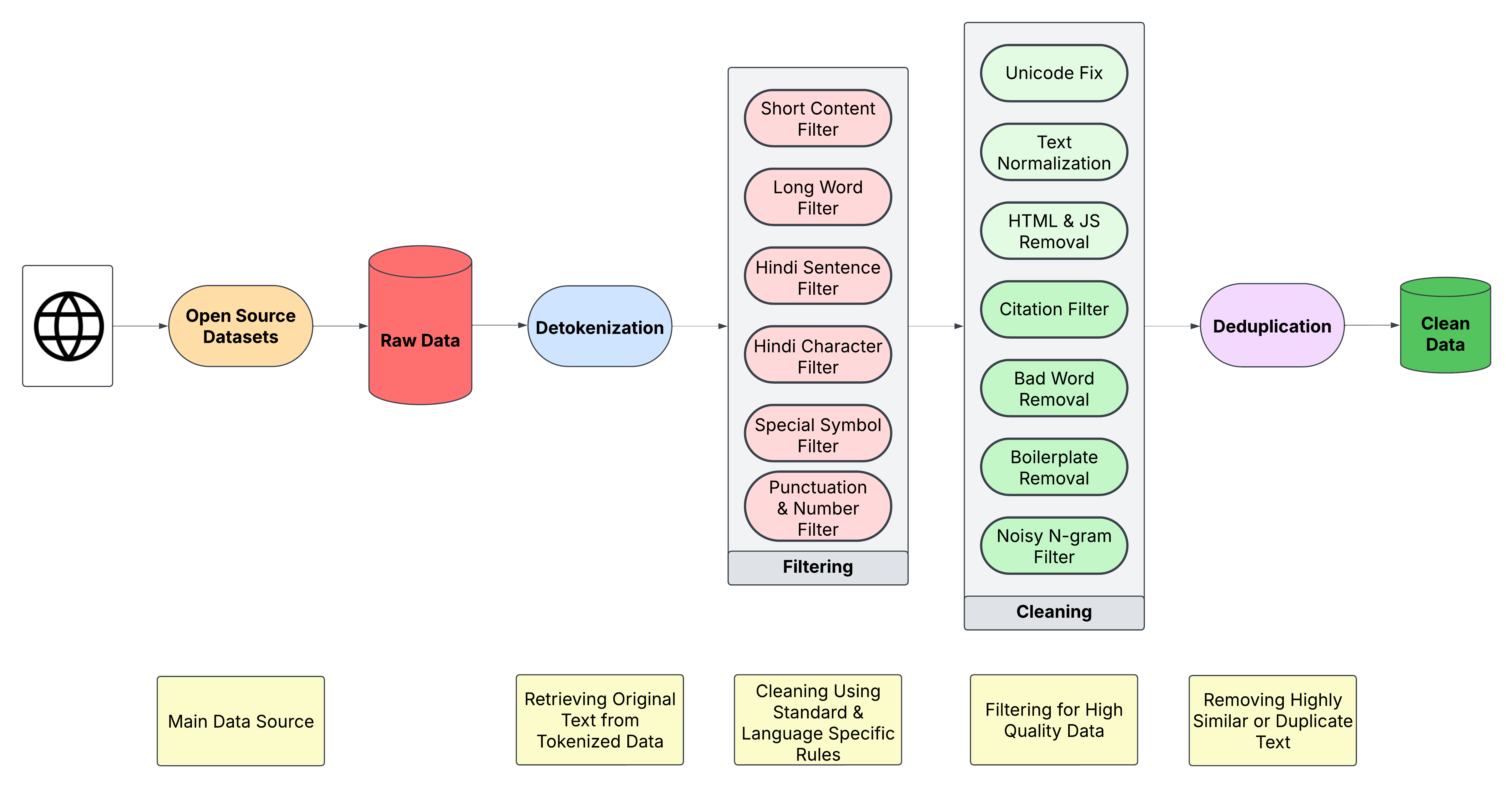}}
    \caption{Our Hindi preprocessing pipeline. We pre-process the raw text through a series of steps. We perform filtering using several heuristics, clean the filtered documents using various techniques, and finally perform deduplication to get the final pretraining corpus.
 }
    \label{fig:preprocessing_hindi}
\end{figure}

We perform a comprehensive pre-processing step on our pre-training data to ensure that \modelname{} learns from diverse, high-quality data.
Here, we provide a brief outline of the workflow of our pre-processing pipeline, which is also illustrated in Figure~\ref{fig:preprocessing_hindi}.

\paragraph{Detokenisation} A large portion of the raw data in our pre-training corpus comes from publicly available datasets, some of which are already pre-processed or tokenised.
To ensure consistency, we \emph{detokenise} the raw data, standardizing the texts across all datasets.
At this stage, all documents in our corpus are non-tokenised regardless of their original source.

\paragraph{Filtering}  Following detokenisation, we filter out irrelevant and low-quality documents using several heuristics as follows: \textit{short content removal}, where documents with less than 20 words are removed; \textit{long word removal}, where documents containing words longer than 100 characters (URLs or gibberish strings) are removed; \textit{Hindi sentence threshold}, where we ensure that at least 50\% of the sentences in each document are in Hindi; \textit{Hindi character threshold}, where we ensure that at least 70\% of the characters in each document are in Hindi; \textit{special symbol removal}, where documents with more than 20\% of their characters as special symbols, punctuation or numerical digits are removed.

\paragraph{Cleaning} We further refine the dataset by cleaning the filtered documents using the following techniques: \textit{Unicode fix}, where we repair corrupted Unicode sequences; \textit{normalization}, where we standardize Hindi punctuation and character forms across the dataset; \textit{HTML/JS removal}, where we remove HTML or JavaScript tags and scripts from each document; \textit{citation removal}, where we remove citations to maintain the text's coherence and logical flow; \textit{boilerplate removal}, where we remove repetitive boilerplate text from each document to reduce redundancy; \textit{bad word removal}, where we detect and remove inappropriate and offensive words / phrases from each document; \textit{noisy n-gram removal}, where we detect and remove meaningless or repetitive n-gram patterns from each document.

\paragraph{Deduplication} 
Finally, we leverage locality-sensitive hashing to perform \emph{deduplication} on the remaining documents. Ultimately, the size of the final pre-processed dataset is reduced to 42\% of the total raw text in the original data sources.

Developing the pre-processing pipeline for Hindi posed greater challenges as compared to English. While English pre-processing pipelines benefit from numerous large-scale, open-access datasets, and well-established techniques, Hindi requires a custom-built approach. Insights gained from experiments with smaller LLMs and the pre-processing pipeline for the pre-training dataset used for \jais{}~\citep{jais} guided the selection of heuristics used in the final pipeline for \modelname{}'s pre-training dataset. However, due to the limited availability of Hindi data, we applied less aggressive filtering than most approaches, ensuring that valuable Hindi content was retained.

\subsection{Mixing Hindi and English Data}

During the adaptation of the Llama-3 model, we mix Hindi and English data following the findings of \citet{gosal2024bilingualadaptationmonolingualfoundation}: continual pre-training of a foundation model on new, previously unseen data can help in bilingual adaptation. 
However, when this new language/domain data is out-of-distribution of the original training data, it can cause forgetting of prior capabilities, which is referred to as a stability gap. 
To mitigate forgetting, we can incorporate a small amount of replay data, closer in distribution to the original pre-training data~\citep{Guo2024StabilityGap}.
In our work, we conduct extensive experiments to determine the minimum proportion of English replay data that should be mixed with Hindi to maintain prior capabilities, while also learning the new language. 

For adapting Llama-3-8B for Hindi, our experiments revealed that a relatively high amount of replay data is necessary.
Specifically, we found that a 1:1 English-to-Hindi dataset mix worked best, enabling cross-lingual capability transfer while preventing saturation of Llama-3 for domain adaptation.
For replay data, we used a mix of textbook, mathematics, coding and reasoning datasets from publicly available sources.

\section{Model}
\label{sec:model}


\subsection{Tokenizer and Architecture} 
\label{sec:model_arc}
\paragraph{\modelname{} Tokenizer} The first step in adapting a foundation model for multilingual use is to construct a balanced vocabulary that includes all target languages. Recent state-of-the-art models such as Llama-3~\citep{Dubey2024TheL3} use byte pair encoding tokenizers~\citep{sennrich-etal-2016-neural}, primarily trained on English data. 

These tokenizers often split non-English words into characters or bytes, creating a significant imbalance among languages. 
This imbalance introduces inefficiency in pretraining, fine-tuning and inference. 
A balanced multilingual tokenizer with low fertility~\citep{rust2021tokenizer} in all languages offers three main advantages: (i) lower training and inference cost; (ii) reduced latency during inference; and (iii) longer context windows~\citep{petrov2023language}. Furthermore, models trained with low-fertility tokenizers tend to perform well on downstream tasks \citep{ahuja-etal-2023-mega}.

\begin{table*}[t]
    \centering
    \small  
    \setlength{\tabcolsep}{6pt}  
    \renewcommand{\arraystretch}{1.2}  
    \begin{tabular}{lcccc} \toprule
        & \textbf{Llama-3} & \textbf{ExtVocab10} & \textbf{ExtVocab20} & \textbf{ExtVocab30} \\ \midrule  
        \textbf{Vocab Size}        & 128,256 & 141,081 & 153,856 & 166,732 \\ \midrule
        \textbf{Hindi Fertility}   & 2.61    & 1.27 (-51.34\%) & \textbf{1.19 (-54.40\%)} & 1.16 (-55.55\%) \\
        \textbf{English Fertility} & 1.35    & 1.35             & 1.35                     & 1.35 \\ \bottomrule
    \end{tabular}
    \caption{Tokenizer intrinsic evaluation across vocab sizes. Adding Hindi vocab reduces fertility by 51.34\%, 54.40\%, and 55.55\% in \textit{ExtVocab10}, \textit{ExtVocab20}, and \textit{ExtVocab30}, respectively, compared to the base Llama-3 tokenizer.}
    \label{tab:vocab_extension}
\end{table*}

In table \ref{tab:vocab_extension}, we show that the Llama-3 tokenizer needs as many as $2.6$ times the number of tokens as words in a given Hindi text. 
Thus, we extend the Llama-3 vocabulary to create a balanced tokenizer for English and Hindi. We add the most frequent Hindi tokens in our pre-training corpora, leading to a larger vocabulary size. We also ensure that the newly introduced tokens are not present in the original vocabulary. We conduct a vocabulary extension analysis to determine the optimal number of new Hindi tokens to be added, ensuring a balanced multilingual vocabulary. The Hindi tokens are borrowed from a monolingual Hindi tokenizer trained on the Hindi corpora. We create a few candidate extended vocabularies and perform intrinsic evaluations following \citet{ali2024tokenizer}.
For intrinsic evaluation, we use the fertility score to measure the efficiency of the tokenization process~\citep{gosal2024bilingualadaptationmonolingualfoundation}. Fertility is defined as \begin{math} f = \frac{S}{W} \end{math}, where \textit{S} is the total number of tokens in the tokenized text and \textit{W} is the number of words in the raw text. It is important to note that fertility is calculated on the held-out subsets of the Hindi corpora, which were not used for tokenizer training. 

 Table \ref{tab:vocab_extension} shows the intrinsic evaluations of three candidate tokenizers, (i) \textbf{\textit{Llama-3-ExtVocab10}},  (ii) \textbf{\textit{Llama-3-ExtVocab20}}, and (iii) \textbf{\textit{Llama-3-ExtVocab30}}, which extend the Llama-3 vocabulary by 10\%, 20\%, and 30\%, respectively.  Based on our tokenizer fertility ablation studies, \textit{Llama-3-ExtVocab20} reduces the fertility of Llama-3’s tokenizer by $54.40$\% while maintaining fertility in English. It achieves a fertility score of $1.19$ on Hindi, which is comparable to the base Llama-3 tokenizer's  English fertility of  $1.35$.  Extending the vocabulary to $30$\% shows minimal improvement in Hindi fertility, therefore, we select \textbf{\textit{Llama-3-ExtVocab20}} as the tokenizer for \modelname{}. 

        
        
        

\paragraph{\modelname{} Embeddings} Following the methods outlined for embedding initialization in \citet{gosal2024bilingualadaptationmonolingualfoundation}, we use a semantic similarity search-based embedding initialization method. This method uses Wechsel multilingual initialization \citep{minixhofer-etal-2022-wechsel} where pretrained embeddings like Fasttext or OpenAI embeddings are used.
For each new Hindi token added to the Llama-3 base vocabulary, we identify top-\( k \) most similar tokens in the base vocabulary based on cosine similarity using embeddings from a pretrained embedding model. We use OpenAI's \texttt{text-embedding-3-large} embeddings \citep{kusupati2024matryoshkarepresentationlearning} for its superior quality and multilingual capabilities. To initialize the embeddings of the new Hindi token, we take a weighted average of the top-\( k \) similar tokens' base embeddings. After experimenting with different values for the \( k \), we achieve the best results with \( k=5 \). This initialization method was used for embedding and unembedding layers of \modelname{}.

\newpage

\paragraph{\modelname{} Architecture} 

Recently, decoder-only models have achieved state-of-the-art performance in generative language tasks.
\modelname{} is derived from Llama-3 8B \citep{Dubey2024TheL3} leveraging the Llama-Pro approach \citep{wu-etal-2024-llama}; hence, it has the standard causal decoder-only  transformer architecture. 
Building upon the Llama-3 model, we incorporated both recent advances from the literature and insights from our own experiments.
Following \citet{wu-etal-2024-llama}, we leverage the block expansion approach, which proves to be highly effective for language adaptation, especially for low-resource languages. By adding and fine-tuning additional decoder blocks initialized to identity mappings while freezing the original Llama-3 backbone, we  train only the newly added blocks. 

This enables the model to integrate new domain and language-specific knowledge without forgetting previously learned information. Although the techniques described in \citet{wu-etal-2024-llama} focus on code and math adaptation, we successfully adapted this approach for language adaptation. We start with Llama-3-8B base model and expanded the number of decoder blocks from $32$ to $40$ using an interleaved approach. A new decoder block was added every $4$ decoder blocks in the base Llama-3 model. In our language adaptation experiments, we found that an optimal data mix of $1:1$ (En:Hi) yielded the best results (in downstream 0 shot tasks in both English and Hindi) compared to Hindi-only adaptation. In both experiments, we trained on a total of $55$B tokens for Hindi in order to maintain the same token count for the appropriate comparison. Our results show that the block-expansion approach is a strong candidate for language adaptation with less training overhead and resources compared to training domain-specific models from scratch, especially for low-resource languages. In the future, this work could expand to other architectures (like Mixture-of-Experts) and modalities, and it would be interesting to analyse the impact on overall accuracy in downstream tasks. Following the results from \citet{gosal2024bilingualadaptationmonolingualfoundation}, we find that the optimal adapter layers are $25$\%  of the existing layers.

\subsection{Pre-Training}

\modelname{} uses a 40-layer architecture with 32 attention heads and a hidden dimensionality of 4096. For optimization, we use a peak learning rate of 1.5e-4 and a batch size of 4 million tokens. For the continual pretraining dataset, we sampled documents from the source list described in Section \ref{sec:pretraining_data} and generated sequences with a context length of 8,192 tokens. When a document was smaller than 8,192 tokens, it was concatenated with other document (documents) and packed into one sequence. \texttt{<|endoftext|>} is used to demarcate the end of each document, giving the language model the information necessary to infer that tokens separated by \texttt{<|endoftext|>} are unrelated.


We train \modelname{} using the AdamW optimizer~\citep{loshchilov2018decoupled} with $\beta_1 = 0.9$, $\beta_2 = 0.95$, $\epsilon = 1e-5$, and weight decay of 0.1. We scale the gradient norms using a maximum norm clipping value of 1.0. The learning rate schedule comprises a linear warm-up to peak learning rate for 1\% of the total steps, followed by a 10$\times$ cosine decay for the rest of the steps. After packing, we used a global batch size of 4M tokens. All training, hyperparameter tuning, and instruction-tuning experiments were conducted on the Condor Galaxy 2 AI supercomputer from Cerebras\footnote{\href{https://www.cerebras.net/blog/introducing-condor-galaxy-1-a-4-exaflop-supercomputer-for-generative-ai/}{*Introducing Condor Galaxy 1: A 4 ExaFLOPS Supercomputer for Generative AI* – Cerebras}} (see Appendix \ref{sec:appendix_training_infrastructure} for details on the training~infrastructure).




\section{Instruction-Tuning}
\label{sec:instruction-tuning}

An effective LLM must accurately interpret user instructions across diverse NLP tasks and adhere to their preferences for helpfulness \& safety.
However, pretraining alone does not enable \modelname{} to accurately interpret and respond to user instructions.
To address this, we instruction-tune~\citep{ouyang2022training} the pre-trained model using a high-quality instruction dataset, aligning the model for practical use-cases and enhancing safety in its responses.


\subsection{Dataset}
\label{sec:ift-data}
\modelname{} is developed as a bilingual model, and thus, it must be enabled to understand instructions in Hindi without compromising its performance in English. 
To this end, we prepare a diverse dataset containing $\sim$81K instructions (Hindi and English) in a prompt-response pair format over a diverse set of NLP tasks including safety-alignment.


\paragraph{\emph{English Instructions}}

The English subset of our instruction-tuning dataset comprises $\sim$39K high-quality instructions spanning a comprehensive range of tasks. 
In particular, we have close to 20K instructions focused on mathematics, while the rest of the examples cover code and various types of reasoning, such as physical, logical and causal reasoning.
Formatted into prompt-response pairs, this subset consists of 7.7M tokens in prompts and 9M tokens in their responses, adding up to a total of $\sim$17M tokens.



\paragraph{\emph{Hindi Instructions}}
As a relatively low-resource language, Hindi does not have many high-quality instruction-tuning datasets. Several existing approaches have utilized machine translation on subsets of English instruction-tuning datasets to create datasets for low-resource languages. 
We create our Hindi instruction-tuning dataset using a similar technique; selecting a set of publicly available English instructions focusing on various forms of reasoning, and translating it into Hindi using various machine-translation models.
We realize that Hindi speakers often use a more relaxed form of the language during informal interactions. We aim for our model to be adept at understanding both formal and informal writing styles. So, we translate the English instructions into two forms of written Hindi:
\begin{itemize}
    \item \textbf{Formal Hindi --} The translated instances are written in Devanagari script with a style of writing consistent with official documents in Hindi.
    \item \textbf{Casual Hindi --} Generated translations contain Hindi (and some English) words using a mix of Devanagari and Latin scripts. This form of the language is generally used by Hindi-speaking individuals during informal conversations like texting, informal speech, interactions on social media, \emph{etc}.
\end{itemize}

Subsequently, several Hindi language experts ensure the quality of translations by manually verifying a sample of instances from the generated dataset.
Ultimately, the Hindi instruction-tuning subset comprises $\sim$22K high-quality machine-translated Hindi instructions, split into $\sim$13.5K in formal Hindi and the remaining in casual Hindi.
In particular, this subset comprises 3.8M prompt tokens and 10M response tokens, or a total of $\sim$14M tokens.

\paragraph{\emph{Safety-Tuning Data}}

We developed a comprehensive safety prompt collection process specifically tailored for Hindi model training, covering eight types of attacks and over 100 detailed safety categories. In the current released version, we randomly sampled 20K data for SFT (see Appendix~\ref{sec:appendix_safety} for more details). 

\subsection{Instruction-Tuning Setup}

As mentioned in Section~\ref{sec:ift-data}, the instances in our raw instruction-tuning data contain a system instruction and a pair of a user-prompt and an AI response. In the case of multi-turn interactions, we have a sequence of multiple prompt--response pairs. Since our model is built on top of \basemodelnamefull{}, we templatize each raw datapoint using the \basemodelnameshort{} prompt template both for supervised fine-tuning (SFT) and for inference.\footnote{\url{https://www.llama.com/docs/model-cards-and-prompt-formats/meta-llama-3/}} At this stage, we oversample the instructions in our dataset (excluding safety instruction-tuning data) to 300\% of the original quantity to strengthen the model. This means we perform SFT over approximately 100M tokens consisting of 47M tokens in Hindi instructions and 53M of the same in English instructions.
Moreover, similar to \jais{}~\citep{jais}, we apply padding to each templatized instance, use the same autoregressive objective as for pretraining, and mask the loss of the prompt to make sure backpropagation considers only the answer tokens during SFT.

\section{Evaluation}
\label{sec:evaluation}

\begin{table}[t]
\centering
\small
\resizebox{\textwidth}{!}{
\begin{tabular}{lcccccc}
\toprule
\multirow{2}{*}{\textbf{Model}} & \multicolumn{1}{c}{\multirow{2}{*}{\textbf{Average}}} & \textbf{MMLU-hi} & \multicolumn{1}{c}{\textbf{HellaSwag-hi}} & \multicolumn{1}{c}{\textbf{ARC-hi}} & \textbf{TruthfulQA-MC1-hi} & \textbf{TruthfulQA-MC2-hi}\\
& & 0-shot & 0-shot & 0-shot & 0-shot & 0-shot \\
\midrule
Airavata-7B & 0.3204 & 0.3044 & 0.3287 & 0.2551 & 0.2600 & 0.4540 \\
Gajendra-v0.1-7B & 0.2949 & 0.3028 & 0.3304 & 0.2594 & 0.2096 & 0.3723 \\
AryaBhatta-GemmaOrca-8.5B & 0.3712 & 0.3682 & 0.4191 & 0.3022 & 0.2962 & 0.4701 \\
AryaBhatta-GemmaUltra-8.5B & 0.3858 & 0.3900 & 0.4394 & 0.3168 & \underline{0.3027} & 0.4801 \\
Nemotron-4-Mini-Hindi-4B-Instruct & \textbf{0.4103} & \underline{0.4294} & \underline{0.4772} & \textbf{0.3579} & \underline{0.3027} & \underline{0.4841} \\
\midrule
Aya-23-8B & 0.3602 & 0.3350 & 0.4481 & 0.2971 & 0.2820 & 0.4390 \\
Mistral-7B-Instruct-v0.3 & 0.3435 & 0.3069 & 0.3435 & 0.2637 & \textbf{0.3053} & \textbf{0.4981} \\
\midrule
Meta-Llama-3-8B & 0.3752 & 0.4010 & 0.4340 & 0.3280 & 0.2630 & 0.4500 \\
Meta-Llama-3-8B-Instruct & 0.3804 & 0.3850 & 0.4070 & 0.3360 & 0.2930 & 0.4810 \\
Llama-3.1-8B-Instruct & 0.3828 & 0.4290 & 0.4500 & 0.3310 & 0.2620 & 0.4420 \\
Llama-3.2-3B-Instruct & 0.3518 & 0.3660 & 0.3860 & 0.2920 & 0.2690 & 0.4460 \\
\midrule
\textbf{Llama-3-Nanda-10B-Chat} & \underline{0.4096} & \textbf{0.4299} & \textbf{0.4922} & \underline{0.3476} & 0.2975 & 0.4810 \\
\bottomrule
\end{tabular}}
\caption{Evaluation results on \textbf{Hindi} benchmarks. \emph{Average} represents the mean score across tasks, and \emph{0-shot} indicates zero-shot results. For all columns, higher the better. \textbf{Bold} represents the best scores in that column while \underline{underlined} represents the second-best scores.}
\label{tab:results:hindi}
\end{table}

In this section, we aim to provide a thorough assessment of the \modelname{} model across a diverse set of evaluation dimensions, covering downstream NLP tasks, safety assessments, and generation capabilities. These evaluations are designed to rigorously measure the model’s performance and adaptability, particularly in supporting multilingual use cases across both Hindi and English languages.

\subsection{Downstream Evaluation}

\paragraph{Evaluation Setup} 
We conduct a comprehensive downstream evaluation, comparing \modelname{} model to a series of baselines that support both Hindi and English languages. Our baseline models include models that are specifically optimized for the Hindi language, such as Gajendra-v0.1~\citep{BhabhaAI_Gajendra_v0.1}, Nemotron-4-Mini-Hindi~\citep{joshi2024}, Airavata~\citep{airavata} and models from the AryaBhatta series~\citep{GenVRadmin1, GenVRadmin2}. We also include multilingual models such as Aya-23~\citep{Aryabumi2024Aya2O} and Mistral~\citep{mistral-7B-v0.3}. Additional models include popular general-purpose models like Llama-3, Llama 3.1, and the latest Llama-3.2~\citep{Dubey2024TheL3}.

We adopt the LM-Evaluation-Harness framework~\citep{eval-harness} to evaluate each model in a zero-shot setting and report the accuracy for each task. Within the framework, the context string is concatenated with each candidate output string, and the answer is determined by selecting the concatenated string with the highest normalized log-likelihood.

\begin{table}[t]
\centering
\small
\resizebox{\textwidth}{!}{
\begin{tabular}{lcccccc}
\toprule
\multirow{2}{*}{\textbf{Model}} & \multicolumn{1}{c}{\multirow{2}{*}{\textbf{Average}}} & \textbf{MMLU} & \multicolumn{1}{c}{\textbf{HellaSwag}} & \multicolumn{1}{c}{\textbf{ARC-en}} & \textbf{TruthfulQA-MC1} & \textbf{TruthfulQA-MC2} \\
& & 0-shot & 0-shot & 0-shot & 0-shot & 0-shot \\
\midrule
Airavata-7B                     & 0.4470 & 0.4044 & 0.6798 & 0.4448 & 0.2607 & 0.4070 \\
Gajendra-v0.1-7B                & 0.4422 & 0.3955 & 0.7308 & 0.4311 & 0.2521 & 0.4062 \\
AryaBhatta-GemmaOrca-8.5B        & 0.5406 & 0.5195 & 0.7370 & 0.4551 & 0.3880 & 0.5406 \\
AryaBhatta-GemmaUltra-8.5B       & 0.5465 & 0.5374 & 0.7573 & 0.4893 & 0.3660 & 0.5465 \\
Nemotron-4-Mini-Hindi-4B-Instruct& 0.5359 & 0.5528 & 0.7122 & 0.4893 & 0.3513 & 0.5021 \\
\midrule
Aya-23-8B                        & 0.4924 & 0.4474 & 0.7431 & 0.4525 & 0.3035 & 0.4924 \\
Mistral-7B-Instruct-v0.3         & \textbf{0.6167} & 0.5898 & \textbf{0.8318} & \textbf{0.5885} & \textbf{0.4211} & 0.5966 \\
\midrule
Meta-Llama-3-8B                  & 0.5526 & 0.6134 & 0.7942 & 0.5338 & 0.2742 & 0.5526 \\
Meta-Llama-3-8B-Instruct         & 0.5911 & 0.6369 & 0.7598 & 0.5689 & 0.3599 & 0.5911 \\
Llama-3.1-8B-Instruct            & 0.5988 & \textbf{0.6644} & 0.7939 & 0.5500 & 0.3696 & 0.5988 \\
Llama-3.2-3B-Instruct            & 0.5338 & 0.5878 & 0.7083 & 0.4577 & 0.3244 & 0.4970 \\
\midrule
\bf Llama-3-Nanda-10B-Chat       & \underline{0.6096} & \underline{0.6499} & \underline{0.8022} & \underline{0.5776} & \underline{0.3995} & \textbf{0.6190} \\
\bottomrule
\end{tabular}}
\caption{Evaluation results on \textbf{English} benchmarks. \emph{Average} represents the mean score across tasks, and \emph{0-shot} indicates zero-shot results. For all columns, higher the better. \textbf{Bold} represents the best scores in that column while \underline{underlined} represents the second-best scores.}
\label{tab:results:english}
\end{table}

We perform the comparative evaluation of \modelname{} against other LLMs for both Hindi and English, building upon the evaluations conducted in prior studies~\citep{Dubey2024TheL3, Aryabumi2024Aya2O, openai2023gpt4}. 

For each language, our evaluation encompasses aspects such as knowledge, reasoning, and misinformation, as outlined in Table~\ref{tab:results:hindi} and Table~\ref{tab:results:english}. 
For Hindi, we assess performance on four translated benchmarks—MMLU-hi, HellaSwag-hi, ARC-hi, and TruthfulQA-[MC1,MC2]-hi that are fetched from Okapi\footnote{\url{https://huggingface.co/alexandrainst}} \citep{dac2023okapi}. For English, following prior studies, we include MMLU~\citep{Hendrycks2020MeasuringMM}, HellaSwag~\citep{Zellers2019HellaSwagCA}, ARC~\citep{Clark2018ThinkYH} and TruthfulQA-[MC1,MC2]~\citep{Lin2021TruthfulQAMH}.


\paragraph{Results for Hindi}
\label{para:res:for:hindi}
Table~\ref{tab:results:hindi} presents the zero-shot evaluation results for Hindi. \modelname{} demonstrates superior performance across many evaluation criteria, placing itself among the state-of-the-art Hindi language models. Specifically, compared to Indic models, such as Gajendra-v0.1, Airavata, AryaBhatta series models, \modelname{} achieves significant absolute improvements across knowledge retrieval, commonsense reasoning and misinformation.
We can further see that among multilingual models, Llama-3.1 and Aya-23-8B are among the best-performing models, with an average accuracy of 38 and 36, respectively. However, \modelname{} outperforms both of them by 2.68 and 4.94 absolute points.
Nemotron-4-Mini-Hindi is the best performing model, outperforming \modelname{} as per average accuracy on log-likelihood evaluations. However, we observe that \modelname{} outperforms Nemotron-4-Mini-Hindi on generation evaluation in Hindi and English (see Section \ref{sec:geneval}) by a significant margin. This highlights the need for comprehensive and more holistic model evaluations to better understand its performance and capabilities.

\paragraph{Results for English}
We also conducted an evaluation for English, with the results shown in Table~\ref{tab:results:english}. Notably, \modelname{} achieves a slight improvement over existing English models. Additionally, we observe that, apart from the AryaBhatta series and Nemotron-4-Mini-Hindi model, other Hindi models, such as Gajendra-v0.1 and Airavata, exhibit significantly lower performance than established English models.
\vspace{-3pt}

\subsection{Generation Evaluation}
\label{sec:geneval}
\begin{figure}[t]
    \centering
    \begin{subfigure}[b]{0.40\textwidth}
        \centering
        \includegraphics[height=2.5cm,width=\textwidth]{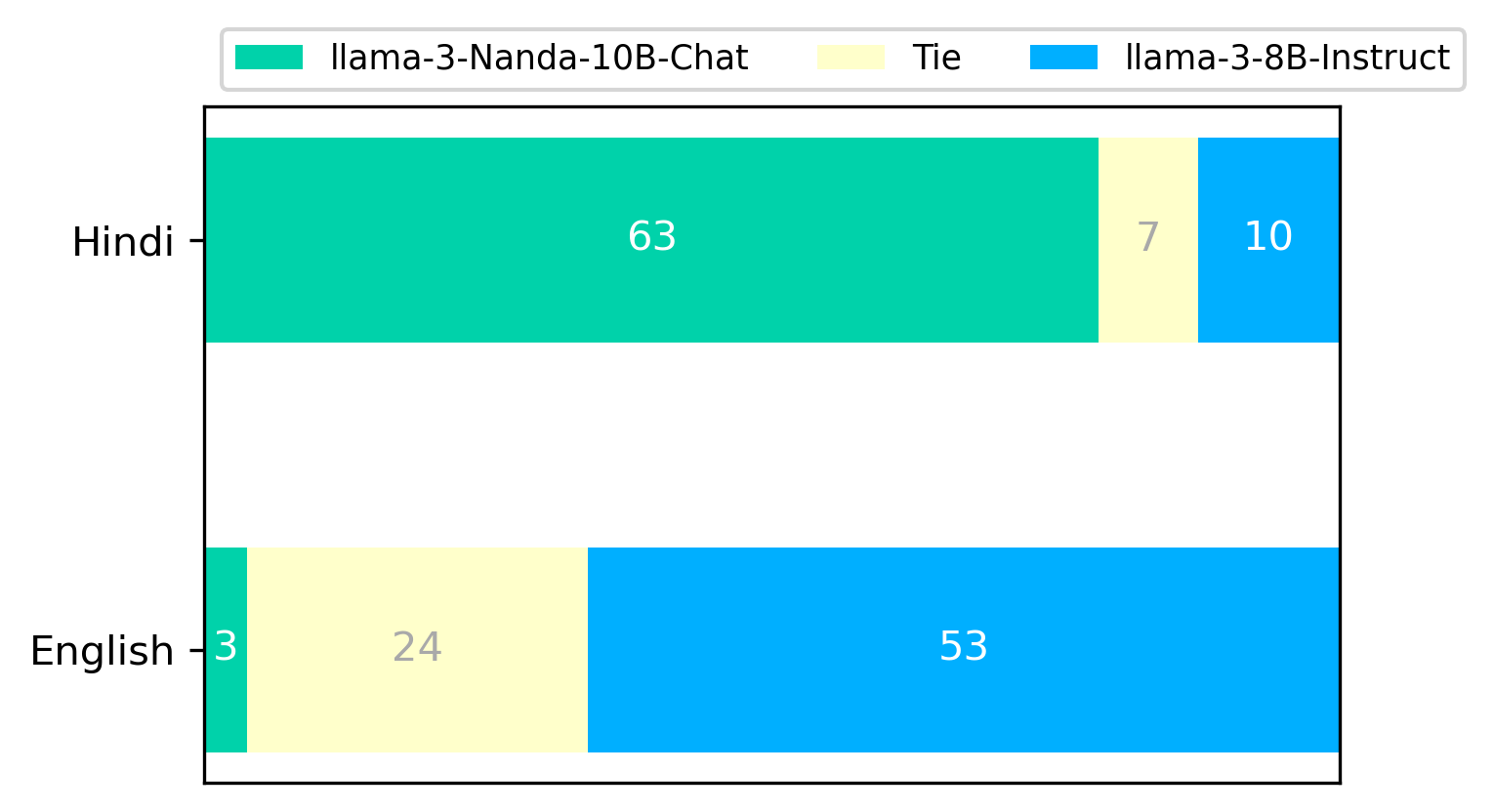} 
        \caption{\small \modelname{} vs Llama-3-8B-Instruct.}
        \label{fig:sub2}
    \end{subfigure}
    \hfill
    \begin{subfigure}[b]{0.40\textwidth}
        \centering
        \includegraphics[height=2.5cm,width=\textwidth]{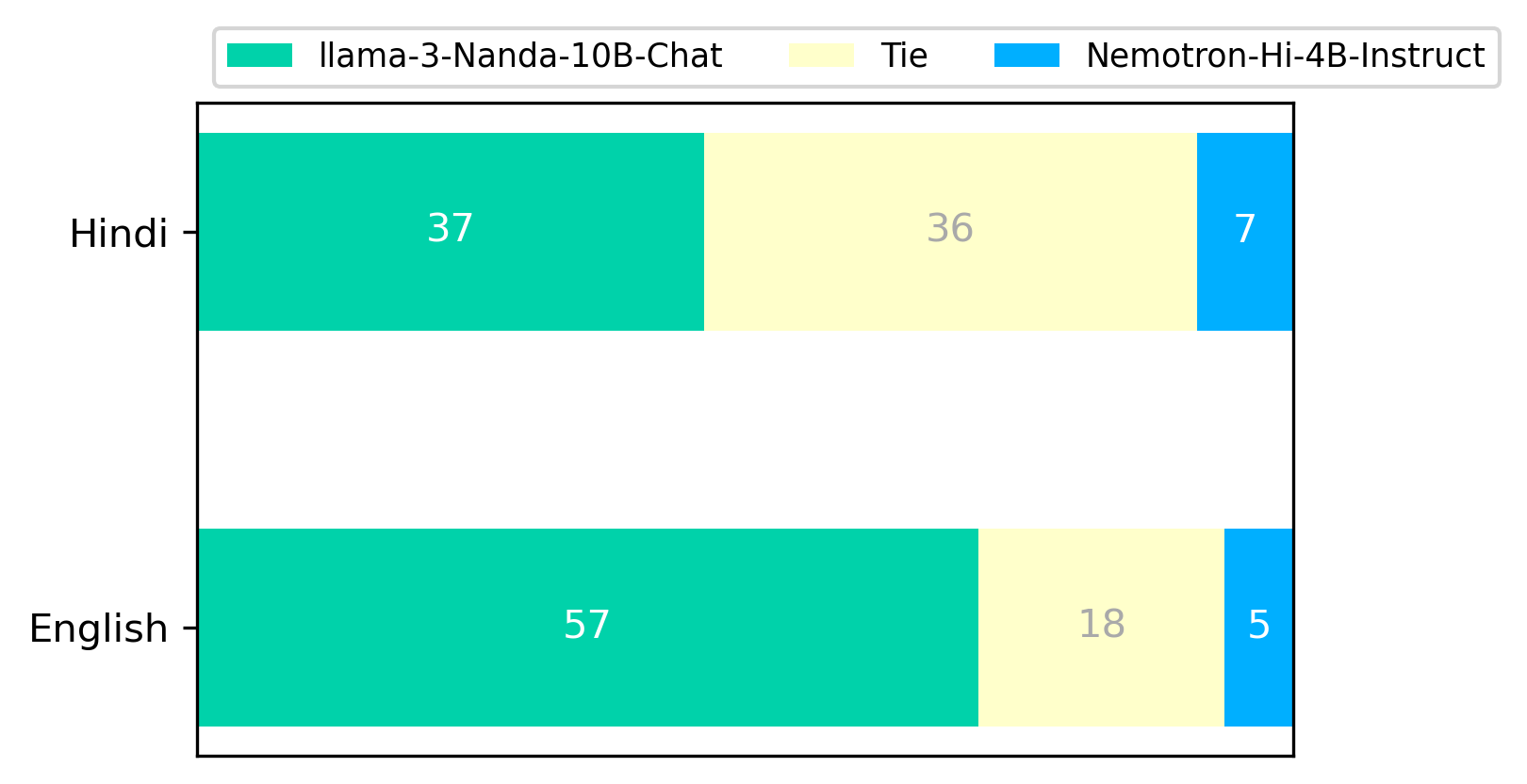} 
        \caption{\small \modelname{} vs Nemotron-Hi-4B-Instruct.}
        \label{fig:sub3}
    \end{subfigure}
    
    \caption{Results for \modelname{} compared to baselines on Vicuna-80 questions, evaluated using GPT-4o as a judge}
    \label{fig:nanda_gen_result}
\end{figure}
In addition to downstream and safety evaluations, we also assess the models’ core capability for Hindi text generation. Consistent with prior studies~\citep{peng2023instruction,vicuna}, we adopt an LLM-as-a-judge evaluation methodology using GPT-4o~\citep{openai2023gpt4}. The evaluation is based on the \textit{Vicuna-Instructions-80}~\citep{vicuna} dataset\footnote{\url{https://lmsys.org/blog/2023-03-30-vicuna/}}, which was manually translated into Hindi by professional translators to ensure linguistic fidelity.

We generate model responses to the Hindi prompts from the \textit{Vicuna-Instructions-80} dataset, using a temperature of 0.3 and a repetition penalty of 1.2. As baselines, we compare against open-source multilingual models such as Llama-3-8B-Instruct~\citep{Dubey2024TheL3} and Nemotron-4-Mini-Hindi-4B-Instruct~\citep{joshi2024} (Nemotron-Hi-4B-Instruct). 

GPT-4o serves as the evaluator, scoring each pair of outputs on a scale from $0$ to $10$ based on quality, relevance, and fluency in Hindi (see Appendix \ref{sec:app:gen_eval_prompt} for our evaluation prompt).

Our generative evaluation results, summarized in Figure~\ref{fig:nanda_gen_result}, show that \modelname{} significantly outperforms all baselines in Hindi text generation. Built upon the Llama-3 (8B) architecture, \modelname{} retains efficiency while introducing improvements that enhance its alignment with the Hindi language, as illustrated in Figure~\ref{fig:nanda_gen_result}:b. Furthermore, \modelname{} surpasses Nemotron-Hi-4B-Instruct, demonstrating superior contextual understanding and generating more natural and fluent Hindi text in language-focused tasks.


\subsection{Safety Evaluation}

\begin{minipage}{0.4\linewidth}
Following previous work~\citep{wang2023dna}, we constructed a novel dataset for Hindi safety evaluation, aiming to identify biases and harmful content within the language model, specifically focused on Hindi and its cultural context. 
The evaluation results from over 1056 risky questions are shown in Table~\ref{tab:results:safety}. We can see that our model achieves similar safety performance to Llama-3.1-8B-Instruct and is much safer than the other models.
\end{minipage}%
\begin{minipage}{0.1\linewidth}
    
\end{minipage}
\hfill
\begin{minipage}{0.5\linewidth}
\vspace{-2.5em}
\centering
\resizebox{\linewidth}{!}{
\small
\begin{tabular}{lcc}
\toprule
\textbf{Model} & \textbf{English} & \textbf{Hindi} \\
\midrule
Airavata-7B & 57.95 & 55.97 \\
Gajendra-v0.1-7B & 44.03 & 39.02 \\
Aya-23-8B & 49.48 & 63.79 \\
AryaBhatta-GemmaOrca-8.5B & 62.88 & 58.14 \\
AryaBhatta-GemmaUltra-8.5B & 61.55 & 50.47\\
Llama-3-8B & 45.45 &  45.46\\
Llama-3.1-8B-Instruct & \bf 90.99 & 87.01 \\
\midrule
\bf Llama-3-Nanda-10B-Chat & 85.97 & \bf 87.96 \\
\bottomrule
\end{tabular}
}
\captionof{table}{Evaluation results for Safety (\% queries where the generated response was safe). \textbf{Bold}
represents the best scores for that language}
\label{tab:results:safety}
\end{minipage}

\vspace{-3pt}

\section{Related Work}
\label{sec:related_work}

Multilingual language models have evolved from English-centric pre-training \citep{devlin2019bert,radford2019,raffel2023,biderman2023} to monolingual models in other languages \citep{faysse2024,Gutiérrez2022,zeng2021,jais,phan2022,koto2020,ko2023} and multilingual training across a few or many languages \citep{nguyen2024,mesham2021,ogueji2021,jude2022,xue2020mt5,chung2023,shliazhko2023,Scao2022BloomA1,lin2022,conneau2020,khanuja2021,oladipo2023,alabi2022,dabre2022}. Models like mT5 \citep{xue2020mt5} and umT5 \citep{chung2023}, trained on the mC4 corpus, offer broad language coverage but primarily rely on unsupervised pre-training and require downstream fine-tuning for specific tasks. Another line of work focuses on expanding language support post hoc through methods such as continued fine-tuning or vocabulary expansion \citep{yong2023,luukkonen2023,lin2024,imani2023}, though these approaches often struggle to scale efficiently. While models such as mBERT \citep{devlin2018bert}, XLM-R \citep{conneau2020}, and Bloom \citep{Scao2022BloomA1} include Hindi, the underrepresentation of Hindi content limits their zero-shot performance relative to monolingual models \citep{li2023bactrianx}.
In contrast to prior work that emphasizes either pre-training or task-specific fine-tuning, our work focuses on enabling instruction-following capabilities in pre-trained multilingual models, allowing them to generalize across tasks without the need for downstream tuning. Appendix~\ref{sec:app:related_work} presents a more comprehensive discussion of related work.


\section{Conclusion}

We have introduced \modelname{}, a new state-of-the-art Hindi-English bilingual instruction-tuned large language model (LLM). It can perform a wide range of generative and downstream language tasks in both Hindi and English, ranging from common-sense reasoning to natural language understanding tasks such as sentiment analysis, irony detection, and hate speech detection. Its pre-trained and fine-tuned capabilities outperform all known open-source Hindi models of similar size and are comparable to state-of-the-art open-source English models that were trained on larger datasets. 

We encourage researchers, hobbyists, and enterprise developers alike to experiment with and develop on top of our model, particularly those working on multi-lingual and/or non-English applications.

\modelname{} represents an important evolution and expansion of the Hindi NLP and AI landscape. This Hindi model, which was born in the UAE, represents an important strategic step for government and commercial organizations towards the digital revolution. By advancing Hindi language understanding and generation, empowering local players with sovereign and private deployment options, and nurturing a vibrant ecosystem of applications and innovation, this work supports a broader strategic initiative of digital and AI transformation to usher in an open, more linguistically inclusive, and culturally-aware era.



\bibliography{colm2025_conference}
\bibliographystyle{colm2025_conference}

\newpage

\appendix

\section{Model Card}
\label{sec:app:model_card}
Table \ref{tab:modelcard} shows the model card~\citep{mitchell2019model} with details about \modelname{}.

\begin{table*}[tbh]
    \centering
    \scalebox{0.80}{
    \begin{tabular*}{\linewidth}{@{\extracolsep{\fill}}p{0.30\linewidth}p{0.70\linewidth}}
   \hline
\multicolumn{2}{c}{\textbf{Model Details}}                                              \\ \hline
\multicolumn{1}{l|}{\textit{Model Developers}}       & {To be released upon acceptance.} \\ \hline
\multicolumn{1}{l|}{\textit{Language(s) (NLP)}}             & {Hindi and English} \\ \hline
\multicolumn{1}{l|}{\textit{Variations}}             & {Instruction-tuned model -- 10B parameters.} \\ \hline
\multicolumn{1}{l|}{\textit{Input}}                  & {Text-only data.} \\ \hline
\multicolumn{1}{l|}{\textit{Output}}                 & {Model generates text.} \\ \hline
\multicolumn{1}{l|}{\textit{Model Architecture}}     & {Llama-3-8B-Base extended by 25\% using the Llama-Pro approach.} \\ \hline
\multicolumn{1}{l|}{\textit{Model Dates}}            & {\modelname{} was trained between June 2024 and September 2024} \\ \hline
\multicolumn{1}{l|}{\textit{Status}}                 & {This static model has been trained using an offline dataset. As we enhance the model safety based on community feedback, upcoming iterations of fine-tuned models will be made available.} \\ \hline
\multicolumn{1}{l|}{\textit{License}}                & {Llama 3} \\ \hline
\multicolumn{2}{c}{\textbf{Intended Use}}                                               \\ \hline
\multicolumn{1}{l|}{\textit{Intended Use Cases}}     & {The \modelname{} 10B model is released with the aim to stimulate research and development in the Hindi NLP community. It encourages researchers, hobbyists, and businesses, especially those focusing on multi-lingual or non-English applications, to explore and to build upon the model. Feedback and collaboration opportunities are welcomed. The model is a pioneering addition to the Hindi LLM ecosystem and has demonstrated exceptional Hindi NLP capabilities compared to other open Hindi or multilingual LLMs globally. Its applications span research advancements in Hindi NLP, and the use of foundational models for fine-tuning.} \\ \hline
\multicolumn{1}{l|}{\textit{Out-of-Scope Uses}}      & {The \modelname{} 10B model is a powerful bilingual Hindi and English language model, but it is important to recognize its limitations and the potential for misuse. Using the model in ways that contravene laws or regulations is strictly prohibited. This encompasses scenarios such as generating or endorsing hate speech, disseminating false information, engaging in illegal activities, managing sensitive data, attempting language generalization beyond Hindi and English, and making critical decisions with high stakes. Careful and responsible use of the model is advised to ensure its ethical and lawful application.} \\ \hline
\multicolumn{2}{c}{\textbf{Hardware and Software}}                                      \\ \hline
\multicolumn{1}{l|}{\textit{Training Factors}}       & {Training was performed on the Condor Galaxy
2 (CG-2) AI supercomputer from Cerebras.} \\ \hline
\multicolumn{2}{c}{\textbf{Training Data}}                                              \\ \hline
\multicolumn{1}{l|}{\textit{Overview}}               & {The training data consists of 65B tokens of Hindi pre-training data along with 21.5M English and 14.5M of Hindi instruction-following tokens.} \\ \hline
\multicolumn{2}{c}{\textbf{Evaluation Results}}                                         \\ \hline
\multicolumn{2}{l}{See downstream, general, and safety evaluation in (Section \ref{sec:evaluation})}                                    \\ \hline
\multicolumn{2}{c}{\textbf{Biases, Risks, and Limitations}}                     \\ \hline
\multicolumn{2}{p{13.6cm}}{{The model is trained on publicly available data, including curated Hindi data, and efforts have been made to reduce unintentional biases in the dataset. However, some biases might still be present, as with all language models. Designed as an AI assistant for Hindi and English, its purpose is to enhance human productivity. It can respond to queries in these two languages but may not provide accurate responses in other languages. Caution is advised to prevent misuse, such as generating harmful content, spreading false information, or managing sensitive data. Responsible and judicious use of the model is strongly encouraged.}}                                    \\ \hline
    \end{tabular*}
    }
\caption{Model card for \modelnamefull{}.}
\label{tab:modelcard}
\end{table*}

\newpage

\section{Generation Evaluation Prompt}
\label{sec:app:gen_eval_prompt}
The prompt provided to GPT-4o for doing the generation evaluation is as follows:
\begin{quote}
\textit{You are a helpful and precise assistant for checking the quality of two Hindi language assistants. Suppose the user speaks only Hindi and Hinglish (Hindi words written in English script), please evaluate both answers with your justification, and provide an integer score ranging from 0 to 10 after your justifications. When evaluating the answers, you should consider the helpfulness, relevance, accuracy, and level of detail of the answers. Do not consider only length as the parameter in level of details, the answer must also be relevant. The score for answer 1 should be wrapped by \texttt{<score1>} and \texttt{</score1>}, and the score for answer 2 should be wrapped by \texttt{<score2>} and \texttt{</score2>}.}
\end{quote}

\section{Training Infrastructure}
\label{sec:appendix_training_infrastructure}
CS-2 systems are purpose-built network-attached AI accelerators. Each CS-2 features 40 GB of SRAM and a peak of 62.5 AI PetaFLOPs, providing a total of 4 ExaFLOPs of AI compute across 64 systems in the CG-2 supercomputer. Utilizing the weight streaming mode of the Cerebras software stack, the Condor Galaxy supercomputers can flexibly schedule multiple jobs based on hardware resource requirements and priority. The number of CS-2s allocated to a job can be dynamically adjusted during training, with performance scaling linearly up to 64 CS-2s per job. This scalability is facilitated by the Cerebras software stack’s use of pure data parallelism to distribute the workload across multiple CS-2s. Jobs are managed by a priority queue system, ensuring efficient allocation of computational resources. 

MemoryX is a large-capacity off-wafer memory service used to store all model weights, gradients, and optimizer states. SwarmX is a broadcast/reduce fabric that connects the memory service MemoryX to each of the CS-2 systems in a wafer-scale cluster. Swarm-X coordinates the broadcast of the model layer weights, giving each CS-2 a local copy, and it receives and aggregates (by addition) the independent weight gradients coming from the CS-2 systems during backpropagation. At the end of each iteration, the aggregated gradients are sent to MemoryX for weight update.

\section{Safety}
\label{sec:appendix_safety}
To ensure high-quality data, a team of five expert annotators initially crafted ``seed prompts'' for direct attack alignment based on previous work by \cite{wang2023dna}, resulting in approximately 1,200 annotated examples focused both on general and Hindi-specific scenarios. Building on this foundation, our expert team guided a 20-member outsourced annotation team, leveraging LLMs, to generate an additional 50K attack prompts, ensuring diversity, linguistic relevance, and thorough coverage for Hindi.

We enrich the set of direct attack prompts in SFT data with a collection of adversarial prompt attack methods. Following~\cite{lin2024achilles}, we adopt eight adversarial prompt attack methods to construct the SFT data. These methods target the following abilities of LLMs: in-context learning, auto-regressiveness, instruction following, and domain transfer, resulting in 100K attack prompts.

To further improve the robustness and generalizability of our model against adversarial prompt attacks, we also adopt LLM-based methods for diversifying the attack prompts. This can also help prevent over-fitting on the attack template used by the works that proposed these attacks. 

Moreover, in the over-refusal prompts task, annotators generate 50K questions that closely resemble potentially unsafe adversarial prompts but are deliberately crafted to be entirely safe. The primary motivation for this task is to address the overrefusal behavior commonly seen in LLMs ~\citep{cui2024orbenchoverrefusalbenchmarklarge}, where models refuse to answer benign questions due to excessive caution. 

By including these prompts, we aim to train the model to better distinguish between genuinely unsafe queries and safe ones, thereby improving the model's responsiveness while maintaining safety.

\paragraph{Taxonomy Development} The development of a detailed taxonomy was the first step in constructing this dataset. This taxonomy categorizes risk areas specific to Hindi, including regional bias, economic situation bias, and national/group character bias. The taxonomy defines specific harms, such as instances of prejudice against particular states in India or negative stereotypes about national characteristics. Example questions were curated to illustrate these biases, helping ensure the evaluation captures a broad range of potential issues.

\paragraph{Data Collection and Translation} The dataset incorporates content sourced in English~\citep{wang2023dna}, initially focused on safety issues like discrimination, toxicity, and adult content, which were then translated into Hindi. The translation process was managed using both automated tools (such as Google Translate and GPT-4) and manual validation by native speakers to ensure the accuracy and cultural relevance of the translations. Each translated entry underwent a thorough validation process to mitigate mistranslations or inadvertent cultural insensitivity.

\paragraph{Annotation and Validation} To ensure the quality of the dataset, we collaborated with outsourced annotators who were provided with guidelines to annotate harmful content. The annotations focus on verifying whether translated content preserved the intended meaning and accurately represented harmful or biased elements in the Hindi context. Annotations were then cross-checked to guarantee consistency and reliability in labelling harmful examples.

\section{Release Notes}
\label{sec:release}

We release \modelname{} under Meta's Llama-3 license, and users must adhere to the terms and conditions of the license,\footnote{\url{https://www.llama.com/llama3/license/}} Meta's acceptable use policy,\footnote{\url{https://www.llama.com/llama3/use-policy/}} Meta's privacy policy,\footnote{\url{https://www.facebook.com/privacy/policy/}} and the applicable policies, laws, and regulations governing the specific use-case and region. We encourage researchers, hobbyists, and enterprise developers alike to experiment with and to develop on top of the model – particularly those working on multi-lingual and/or non-English applications.

\subsection{Intended Use}

This model is one of the first of its kind in the Hindi LLM ecosystem and has shown to be the best in the world among open Hindi or multilingual LLMs in terms of Hindi NLP capabilities. Some potential downstream uses are listed below:

\begin{itemize}
 \item Research: This model can be used by researchers and developers to advance the Hindi LLM/NLP field.
 \item Commercial Use: It can be used as a foundational model to further fine-tune for specific use cases. Some potential use cases for businesses include (1) chat assistants, (2) downstream tasks such as NLU/NLG, (3) customer service, and (4) process automation.
\end{itemize}

We believe that a number of audiences will benefit from our model:

\begin{itemize}
\item Academics: those researching Hindi natural language processing.
\item Businesses: companies targeting Hindi-speaking audiences.
\item Developers: those integrating Hindi language capabilities in apps.
\end{itemize}

\subsection{Out-of-Scope Use}

While \modelname{} is a powerful bilingual model catering to Hindi and English, it is essential to understand its limitations and the potential for its misuse. The following are some examples from the long list of scenarios where the model should not be used:

\begin{itemize}

\item \textbf{Malicious Use}: The model should not be used for generating harmful, misleading, or inappropriate content. This includes but is not limited to (\emph{i})~generating or promoting hate speech, violence, or discrimination, (\emph{ii})~spreading misinformation or fake news, (\emph{iii})~engaging in illegal activities or promoting them, (\emph{i})~(\emph{iv})~handling sensitive information: the model should not be used to handle or to generate personal, confidential, or sensitive information.

\item \textbf{Generalization Across All Languages}: \modelname{} is bilingual and optimized only for Hindi and English. It should not be assumed to have equal proficiency in other languages or dialects.

\item \textbf{High-Stakes Decisions}: The model should not be used for making high-stakes decisions without human oversight. This includes medical, legal, financial, or safety-critical decisions, among others.
\end{itemize}

\subsection{Biases, Risks, and Limitations}
The model is trained on a mix of publicly available and proprietary data, which in part was curated by our preprocessing pipeline. We used different techniques to reduce the bias that is inadvertently present in the dataset. While efforts were made to minimize biases, it is still possible that our model, like all LLM models, may exhibit some biases.

The model is trained as an AI assistant for Hindi and English speakers, and thus, it should be used to help humans boost their productivity. In this context, it is limited to producing responses for queries in these two languages, and it might not produce appropriate responses for queries in other languages.

Potential misuses include generating harmful content, spreading misinformation, or handling sensitive information. Users are urged to use the model responsibly and with discretion.

\section{Additional Related Work}
\label{sec:app:related_work}

Below, we discuss some more previous work on the following relevant topics: LLMs in general, multilingual models, instruction-tuning, and evaluation of LLMs.

\paragraph{Multilingual Models}

Pre-training a language model typically involves using unsupervised learning with large datasets. While much of this work has been centered on English \citep{devlin2019bert,radford2019,raffel2023,biderman2023}, significant research has also been dedicated to mono-lingual pre-training in languages other than English \citep{faysse2024,Gutiérrez2022,zeng2021,jais,phan2022,koto2020,ko2023}, as well as training models on a small number of languages \citep{nguyen2024,mesham2021,ogueji2021,jude2022}. 

There have also been massively multilingual pre-training efforts \citep{xue2020mt5,chung2023,shliazhko2023,Scao2022BloomA1,lin2022,devlin2019bert,conneau2020,khanuja2021,oladipo2023,alabi2022,dabre2022}. Models based on the mC4 corpus \citep{xue2020mt5}, which cover approximately 100 languages, represent the broadest range of coverage in pre-trained models available today. Notable examples include mT5 \citep{xue2020mt5} and umT5 \citep{chung2023}, which are the largest publicly accessible multilingual pre-trained models. 

However, a key limitation of all these approaches is that they focus on pre-training, requiring users to perform downstream task fine-tuning for specific applications. In contrast, our work emphasizes equipping pre-trained models with instruction-following capabilities.

Another important research direction focuses on adapting pre-trained models to accommodate new languages not included during the initial training phase. These studies explore methods such as continued fine-tuning and embedding space adaptation. For instance, previous work \citep{yong2023,luukkonen2023} has expanded language coverage by gradually adding languages through additional pre-training on monolingual datasets, a method that does not scale efficiently. In a concurrent effort, \citep{lin2024} extends language coverage significantly by using vocabulary expansion and further pre-training Llama-2 with Glot500-c \citep{imani2023}. 

Hindi has also been integrated into these multilingual models, including earlier models such as mBERT \citep{devlin2018bert} and XLM-RoBERTa \citep{conneau2020}, as well as more recent large language models such as Bloom \citep{Scao2022BloomA1}. However, due to the Hindi content being dwarfed by other languages, these models tend to perform substantially worse than dedicated monolingual models and often exhibit limited generalization abilities in zero-shot settings \citep{li2023bactrianx}.




\paragraph{Evaluating Large Language Models}

Large language models are highly capable of generating coherent and fluent text but often struggle with factual accuracy and reasoning abilities. To assess factual accuracy, models like GPT-4 \citep{openai2023gpt4} and Llama \citep{touvron2023llama} use school exam-style questions \citep{hendrycksmeasuring} to gauge how faithfully they can provide knowledge. Common-sense reasoning is also critical and is tested through datasets such as \emph{HellaSwag}~\citep{Zellers2019HellaSwagCA}, \emph{WinoGrande}~\citep{sakaguchi2021winogrande}, \emph{ARC} easy and challenge~\citep{clark2018think}, and \emph{OpenBookQA}~\citep{mihaylov-etal-2018-suit}. For evaluating reasoning through programming, benchmarks like HumanEval~\citep{chen2021evaluating} and MBPP~\citep{austin2021program} are used.

In the domain of Hindi NLP, \cite{kakwani2020} introduced IndicGLUE, the first Indic NLU benchmark for 11 languages, while \cite{doddapaneni2023} expanded upon this by releasing IndicXTREME, covering all 22 Indic languages. On the natural language generation (NLG) side, \cite{kumar2022} developed the IndicNLGsuite, which supports five tasks across 11 languages. Additionally, \cite{gala2023} presented IN22, a machine translation benchmark for evaluating both conversational and general translation across all 22 languages. More recently, \cite{singh2024} proposed IndicGenBench, a benchmark covering diverse tasks such as cross-lingual summarization, machine translation, and cross-lingual question answering. \cite{watts2024} evaluated models using LLMs and humans and observed that they agree fairly well on most Indic languages.



\end{document}